\documentclass[journal]{IEEEtran}

\usepackage{color}
\usepackage[pdftex]{graphicx}

\usepackage[cmex10]{amsmath}
\usepackage{array}
\usepackage{mdwmath}
\usepackage{mdwtab}
\usepackage{eqparbox}
\usepackage{url}
\usepackage{cite}
\usepackage{verbatim}
\usepackage{multirow}
\usepackage{amsmath}
\usepackage{booktabs}

\begin{document}

\title{Multi-organ Segmentation over Partially Labeled Datasets with Multi-scale Feature Abstraction}

\author{Xi Fang, Pingkun Yan,~\IEEEmembership{Senior Member, IEEE}
\thanks{Asterisk indicates the corresponding author.}%
\thanks{X.~Fang and *P.~Yan are with the Department of Biomedical Engineering and the Center for Biotechnology and Interdisciplinary Studies at Rensselaer Polytechnic Institute, Troy, NY, USA 12180. (e-mail: fangx2@rpi.edu, yanp2@rpi.edu).}%
\thanks{This work was partially supported by National Institute of Biomedical Imaging and Bioengineering (NIBIB) of the National Institutes of Health (NIH) under awards R21EB028001 and R01EB027898, and through an NIH Bench-to-Bedside award made possible by the National Cancer Institute.}
}

\maketitle

\makeatletter
\def\ps@IEEEtitlepagestyle{
	\def\@oddfoot{\mycopyrightnotice}
	\def\@evenfoot{}
}
\def\mycopyrightnotice{
	{\footnotesize
		\begin{minipage}{\textwidth}
			\centering
			Copyright~\copyright~2020 IEEE. Personal use of this material is permitted. However, permission to use this  \\ 
			material for any other purposes must be obtained from the IEEE by sending a request to pubs-permissions@ieee.org.
		\end{minipage}
	}
}

\begin{abstract}
Shortage of fully annotated datasets has been a limiting factor in developing deep learning based image segmentation algorithms and the problem becomes more pronounced in multi-organ segmentation. In this paper, we propose a unified training strategy that enables a novel multi-scale deep neural network to be trained on multiple partially labeled datasets for multi-organ segmentation. In addition, a new network architecture for multi-scale feature abstraction is proposed to integrate pyramid input and feature analysis into a U-shape pyramid structure. To bridge the semantic gap caused by directly merging features from different scales, an equal convolutional depth mechanism is introduced. Furthermore, we employ a deep supervision mechanism to refine the outputs in different scales. To fully leverage the segmentation features from all the scales, we design an adaptive weighting layer to fuse the outputs in an automatic fashion. All these mechanisms together are integrated into a Pyramid Input Pyramid Output Feature Abstraction Network (PIPO-FAN). 
Our proposed method was evaluated on four publicly available datasets, including BTCV, LiTS, KiTS and Spleen, where very promising performance has been achieved. The source code of this work is publicly shared at \url{https://github.com/DIAL-RPI/PIPO-FAN} to facilitate others to reproduce the work and build their own models using the introduced mechanisms.
\end{abstract}

\begin{IEEEkeywords}
Medical image segmentation, multi-scale feature, deep learning, convolutional neural networks, multi-organ segmentation, multiple datasets
\end{IEEEkeywords}

\IEEEpeerreviewmaketitle


\section{Introduction}

\IEEEPARstart{A}{utomatic} multi-organ segmentation, an essential component of medical image analysis, plays an important role in computer-aided diagnosis. For example, locating and segmenting the abdominal anatomy of CT images can be very helpful in cancer diagnosis and treatment \cite{rueckert_atlas_seg_2013}. 
With the surge of deep learning in the past several years, many deep convolutional neural network (CNN) based methods have been proposed and applied to medical image segmentation \cite{li_hdenseu, zhu_boundary, 8451958, DR_drinet}. 
Two main strategies to improve image segmentation performance are: (i) Designing better model architectures and (ii) Learning with larger scale of labeled data.

\begin{figure}[t]
	\centering
\includegraphics[width=3.5in]{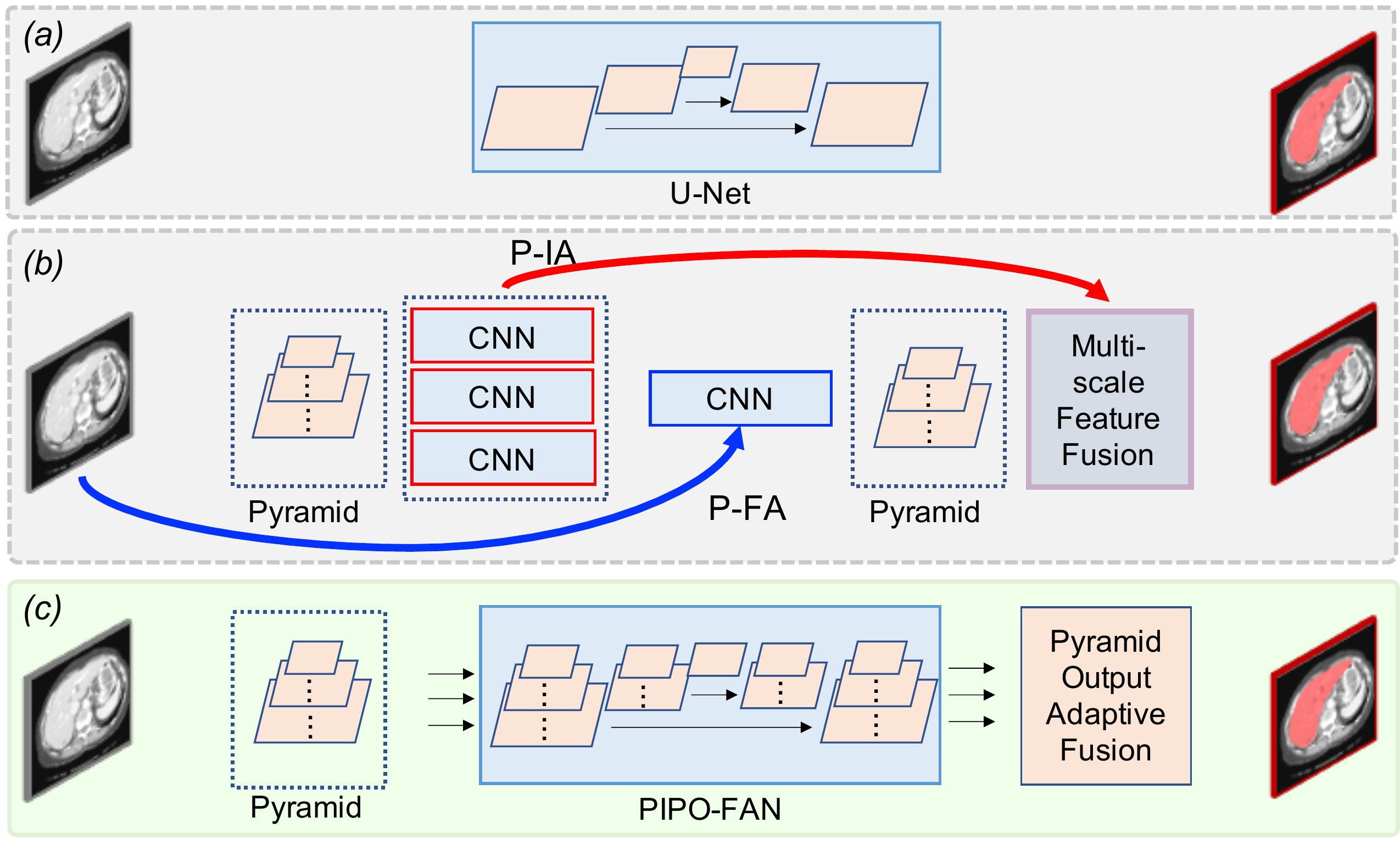}
\caption{
(a) Pyramid convolution structure of Skip-Net in the most commonly seen form of U-Net. (b) Pyramid input analysis (P-IA) applies pyramid parsing to the input and pyramid feature analysis (P-FA) uses pyramid parsing on the intermediate features. (c) Our proposed pyramid feature abstraction network analyzes pyramid input with Equal Depth Convolution (EDC) and fuses the pyramid output to achieve improved segmentation.}
\label{fig:multiscale}
\end{figure}

The state-of-the-art models in medical image segmentation are variants of the encoder-decoder architecture, such as fully convolutional network (FCN) \cite{LongSD15} and U-Net \cite{unet}. A major focus of the FCN based segmentation methods has been on network structure engineering by incorporating multi-scale features. That is because multi-scale features contain detailed texture information combined with contextual information, which are beneficial for semantic image segmentation. The existing deep learning image segmentation methods that exploit multi-scale features generally come with designs of pyramid structure using skip connections or pyramid parsing modules. Networks employing skip connections to exploit features from different levels are referred as skip-nets \cite{Chen2015AttentionTS}. Features in skip-net are multi-scale in nature due to the increasing size of receptive field. U-Net \cite{unet} as shown in Fig.~\ref{fig:multiscale}(a) is a typical skip-net with pyramid structure, which is commonly used as a baseline network to learn pixel-wise information in medical image segmentation. Many works improve segmentation performance on top of U-Net by incorporating new convolutional blocks such as residual blocks \cite{He2016DeepRL} and dense blocks \cite{Huang2017DenselyCC}. For instance, Han~\cite{han_automatic_2017} won ISBI 2017 LiTS Challenge\footnote{https://competitions.codalab.org/competitions/15595} by replacing the convolutional layers in U-net with residual blocks from ResNet. Li et al. \cite{li_hdenseu} replaced the encoder part of U-net with DenseNet-169 and obtained a high accuracy of Dice 95.3\% on liver CT segmentation.  

In addition to exploring new convolutions in pyramid network structures to efficiently extract high level features, incorporating pyramid parsing in FCN also helps utilize multi-scale information in segmentation tasks \cite{lin2016efficient,psp,fang2019unified}. Fig.~\ref{fig:multiscale}(b) illustrates two general kinds of pyramid parsing modules, pyramid input analysis (P-IA) and pyramid feature analysis (P-FA). P-IA applies CNN on input pyramid to extract multi-scale features through parallel convolutional channels \cite{lin2016efficient,kamnitsas2017efficient}. P-FA methods perform pyramid parsing after the features have been extracted by a CNN for further abstraction. In those works, features from different scales are only combined at very late stage of the networks to generate final output labels \cite{psp}.
	
In our work, we hypothesize that extracting and maintaining multi-scale features through the network to glean hierarchical contextual information can significantly improve the segmentation performance. Although pyramid parsing modules and pyramid structures with skip connections have been utilized in various computer vision tasks (e.g. saliency detection, object detection, image segmentation), these mechanisms have not been integrated together for exploration, especially in multi-organ segmentation from medical images.
In this paper, as shown in Fig.~\ref{fig:multiscale}(c), we design a new network architecture, which takes pyramid inputs with dedicated convolutional paths to combine features from different scales to utilize the hierarchical information. The hierarchical convolutions through different scales alleviate the semantic gaps between ends of connections. 
To fuse the segmentation features from different scales, we further design an adaptive weight layer. In particular, this layer uses an attention mechanism to compute the importance of the features from each scale. The proposed method is thus coined as Pyramid Input and Pyramid Output (PIPO) Feature Abstraction Network (FAN). The proposed method can be easily integrated into other existing U-shape networks to improve the feature representation power of the models.

\begin{figure*}[t]    
	\center{\includegraphics[width = \textwidth] {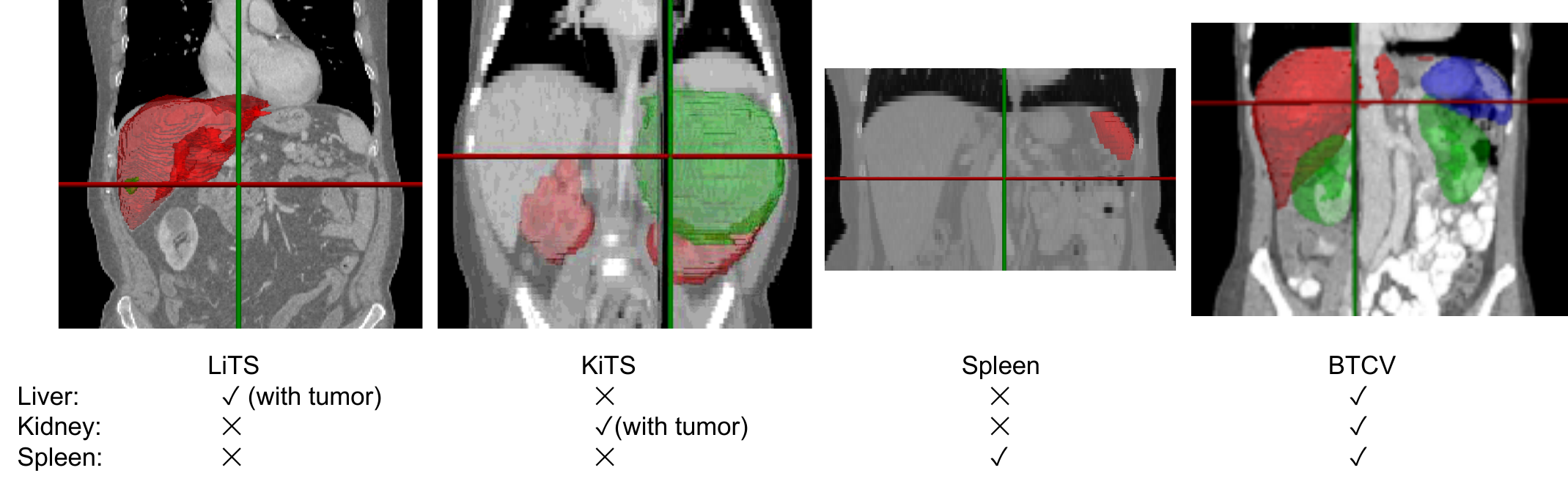}}        
	\caption{Sample images from the datasets of LiTS, KiTS, Spleen, and BTCV. Some organs are included in all the datasets but only annotated in no more than two of them. 
	}
	\label{fig:data_examples}
\end{figure*}

Deep CNNs have shown great performance for single organ segmentation. However, we often face the problem of multi-organ segmentation in clinical applications. Segmenting multiple organs independently using single organ segmentation algorithms may be straightforward, which, however, loses the holistic view of the image. Therefore, the segmentation performance may be degraded. However, collecting multi-organ annotations for training algorithms is more difficult than annotating single organ datasets. Ideally, researchers could use similar datasets created by different organizations for their research. However, in reality, there will always be some differences that the datasets cannot be directly used, since those data were collected for various purposes.
For instance, there are several abdominal CT datasets are publicly available but they are annotated with different targeted organs at risk as shown in Fig.~\ref{fig:data_examples}. One CT dataset contains the labeled segmentation of the spleen, while another dataset include only the liver annotation. 
It will be highly advantageous if we can utilize all those datasets together to train a multi-organ segmentation network. In order to achieve this goal, in this paper, we propose a unified training strategy with a novel target adaptive loss.

In this paper, we extensively evaluated the proposed method on the BTCV (Beyond  the  Cranial  Vault)  segmentation challenge dataset\footnote{https://www.synapse.org/\#!Synapse:syn3193805/wiki/89480} and three partially labeled datasets, including MICCAI 2017 Liver Tumor Segmentation (LiTS) Challenge\footnote{https://competitions.codalab.org/competitions/17094} dataset, MICCAI 2019 Kidney Tumor segmentation (KiTS) Challenge \footnote{https://kits19.grand-challenge.org/} dataset, and the spleen segmentation dataset\cite{simpson2015chemotherapy}. We demonstrate very promising performance of medical image segmentation on these datasets compared to the state-of-the-art approaches.

Our contributions in this work can be summarized as follows. 
\begin{enumerate}
\item 
A new pyramid-input and pyramid-output network is introduced to condense multi-scale features to reduce the semantic gaps between features from different scales.

\item
An image context based adaptive weight layer is used to fuse the segmentation features from multiple scales.

\item
A target adaptive loss is integrated with a unified training strategy to enable image segmentation over multiple partially labeled datasets with a single model. 

\item
Very competitive performance with state-of-the-arts has been achieved by using the developed network on multiple publicly available datasets.
\end{enumerate}

\section{Related Works}

\subsection{Multi-scale feature learning}

Multi-scale features contain detailed texture and context information, which is helpful for many computer vision tasks including object detection \cite{liu2020deep}, saliency detection \cite{dong2018holistic} and image segmentation \cite{xueying_isbi}. Multi-scale feature learning can be generally grouped into two categories. The first type, sometimes referred to as skip-net \cite{Chen2015AttentionTS}, combines different level features with skip connections. They are always in pyramid structures. For example, FPN \cite{lin2017feature}, U-Net \cite{unet} and FED-Net \cite{oktay2018attention} use an encoder that gradually down-samples to capture more context, followed by a decoder that learns to upsample the segmentation. Low-level fine appearance information is fused into coarse high-level features through skip connections, attention gates or convolutional blocks between shallow and deep layers. 
These works effectively fuse multi-scale context using skip connections, but in the same time introduces huge semantic gap between features at two ends of the connections. UNet++ \cite{10.1007/978-3-030-00889-5_1} tries to bridge the semantic gap of skip-net by redesigning the skip-pathways to fuse semantic similar features.
Pyramid structures have been employed to extract multi-scale features in computer vision tasks. 
To obtain effective pyramid features, deep supervision has been used in saliency detection \cite{dong2018holistic} and image segmentation \cite{deepsup}. In object detection works using pyramid features \cite{lin2017feature, liu2020deep}, detection is made at different levels in the feature pyramid for objects of various sizes.
Pyramid features are fused to provide multi-scale context for the final prediction. The mechanism has been shown to be highly useful, which, however, has not been explored in multi-organ segmentation.
SegCaps \cite{lalonde2018capsules} introduces Capsules for object segmentation, which replaces max-pooling layers with convolutional strides and dynamic routing to preserve spatial information. That achieves highly competitive segmentation performance compared with UNet, but with substantially decreased parameter space.

The second type of methods uses pyramid parsing module to  extract multi-scale features in the same convolutional level with either pyramid input analysis (P-IA) or pyramid feature analysis (P-FA) as shown in Fig.~\ref{fig:multiscale}. These features have different effective receptive fields and are concatenated or summed to boost feature representation ability of context information. 
For example, P-FA methods like PSP-Net \cite{psp} apply spatial pyramid pooling to convolutional feature maps for pyramid feature analysis. 
Deeplab \cite{DBLP:journals/pami/ChenPKMY18} and CE-Net \cite{8662594} use parallel atrous convolution with different sampling rates to extract multi-scale features to augment segmentation. Qin et~al. \cite{qin2018autofocus} integrates attention module into the pyramid parsing layer to adapt the network's receptive field in a data-driven manner.
P-IA methods, on the other hand, perform feature analysis from input images with different sizes, i.e. create image pyramid at multiple scales. 
For example, Farabet et~al. \cite{hier_lecun} enforce scale invariance by applying shared network on different scales of a Laplacian pyramid version of the input image. Kamnitsas et~al. \cite{kamnitsas2017efficient} employ a dual pathway architecture that processes the input images at multiple scales simultaneously to extract pyramid features to strengthen the feature representation.

Although pyramid inputs and pyramid parsing module are commonly utilized in computer vision, they have not been explored in multi-organ segmentation. Since both the pyramid parsing module and skip-net can extract multi-scale context information to help image segmentation, they may also be combined to further boost the performance. Some recent works like \cite{mimo_net, fu2018joint} integrate features from pyramid input images to the U-Net structure. However, since those features are at different semantic abstraction levels, fusing those features from different scales may cause the problem of semantic gap. Thus, those networks fail to mitigate the multi-scale context information, by only partially utilizing the pyramid shape of U-Net.

\begin{figure*}[t]   
	\center{\includegraphics[width = \textwidth] {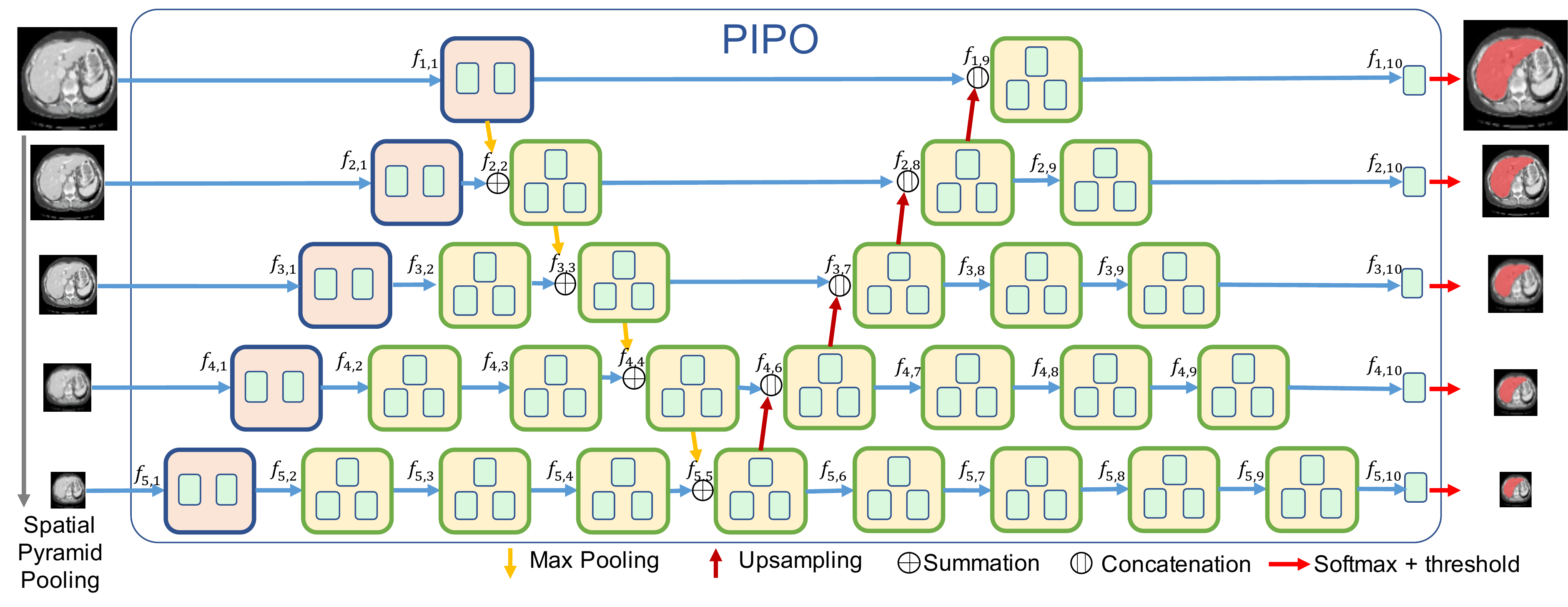}}        
	\caption{Overview of the PIPO architecture. With the designed architecture, image information propagates from pyramid input to pyramid output through hierarchical abstraction and combination at each level.} 
	\label{fig:Concave}
\end{figure*}

\subsection{Segmentation over multiple datasets}

Various datasets for semantic segmentation have been presented. Training one general model over multiple datasets will make the features more robust and accurate. 
When annotations for different datasets are same labels \cite{rundo2019use}, model can be directly trained over these datasts \cite{peng2019method}. However, in most cases, different datasets have different annotations.
Although many datasets share similar appearance information, annotations difference make it a challenging problem of model generalization over multiple datasets.  
Many works have been done to deal with the diversity problem. These works can be divided according to different types of annotation difference. When strongly label (pixel-wise annotations) and  weak labels (image category, bounding box) exists over multiple datasets. It's a problem of semi-supervised segmentation. Features are extracted from the encoder are used to learn through multi-task learning.
Hong et~al. \cite{hong2016learning} train class labels and segmentation together with two branches. 
Papandreou et.al. \cite{papandreou2015weakly} develop Expectation-Maximization (EM) methods for semantic image segmentation model training on few strongly labeled and many weakly labeled images, sourced from one or multiple datasets.
When the labels are different but the types of annotations are pixel-wise. Only a few works have been done to solve the problem. Some works \cite{meletis_heter_2018, kong_heseg_2019} design hierarchical classifier for multiple heterogeneous datasets. Each classifier classifies the children labels of a node and the whole classifier is trained. However, semantic hierarchy of the labels is required. The Multi-Sourced Dice Loss proposed by Tang et al. \cite{tang2019improving} is another closely related work, which was used to train a segmentation network on heterogeneous multi-resource datasets to segment the spleen. Unlike these methods, our proposed approach allows a single model using partial labels, which exploit label proportion information.  To train our model, we introduce a new loss function that adapts it-self to the proportion of known labels per example.

\subsection{Multi-organ Segmentation}

Accurate and robust segmentation of multiple organs is essential. Three methodologies, statistical models\cite{cerrolaza2015automatic, okada2015abdominal}, multi-atlas methods \cite{robin_2012_miccai, tong2015discriminative} and registration-free methods \cite{herve_2014_miccai, zografos2015hierarchical} are always used to do multi-organ segmentation. However, these methods are always organ-specific and require prior professional knowledge and manual designing.

Recent advances in deep learning and data availability enabled the training of more complex registration-free methods, eg. deep CNN, which neither explicit anatomical correspondences nor hand crafted features. Many studies based on deep CNNs focused on single organ segmentation, particularly for abdominal regions due to the similar intensity and size variation between different target organs.
Multi-organ segmentation in abdominal CT has been an important problem to solve for precise diagnosis and treatment. Deep learning based segmentation methods have been developed to segment multiple abdominal organs \cite{eli_2018_deepvnet, peng2019method, chen2017towards, roth2018application, wang2019abdominal}. Some works use two-step segmentation to exploit the prior anatomical information \cite{chen2017towards, roth2018application, wang2019abdominal}. Specifically, Chen et~al. \cite{chen2017towards} use an organ attention module to guide the fine segmentation. However, majority of the existing works are mostly customized for a certain dataset labeled with all target organs. The lack of variety and quality of datasets make segmentation models specific to particular diseases and also hard to train.
 
Relaxing the learning requirements to exploit all the available labels open better opportunities for creating large-scale datasets for training deep neural networks. A promising strategy is to use partial annotations from multiple available datasets, which may shared some labeled organs in common. Partial labels are recently introduced to improve image classification performance \cite{Durand_2019_CVPR}.  
We suppose that different datasets are complementary, which can be used together to train a unified segmentation model without harming the performance.

\section{Pyramid Input and Pyramid Output Feature Abstraction Network}

In this section, we present a novel Pyramid Input and Pyramid Output Feature Abstraction Network (PIPO-FAN), which fully fuses multi-scale context information and semantic similar features with one single network. Pyramid input analysis and pyramid feature analysis are integrated in the proposed network. 
Our hypothesis is that the semantic information in various depths can be further enhanced by utilizing hierarchical contextual features. PIPO-FAN aims to effectively extract multi-scale features for medical image segmentation, on top of the multi-scale nature of U-net.
Fig.~\ref{fig:Concave} shows the overall structure of the proposed PIPO. The network perform spatial pyramid pooling on input and hierarchical abstract multi-scale features at each level enforced by deep supervision mechanism.

\subsection{Pyramid Input with Equal Convolutional Depth}
\label{sec:ECD}

\begin{figure*}[t]
\center{\includegraphics[width=\textwidth]
		{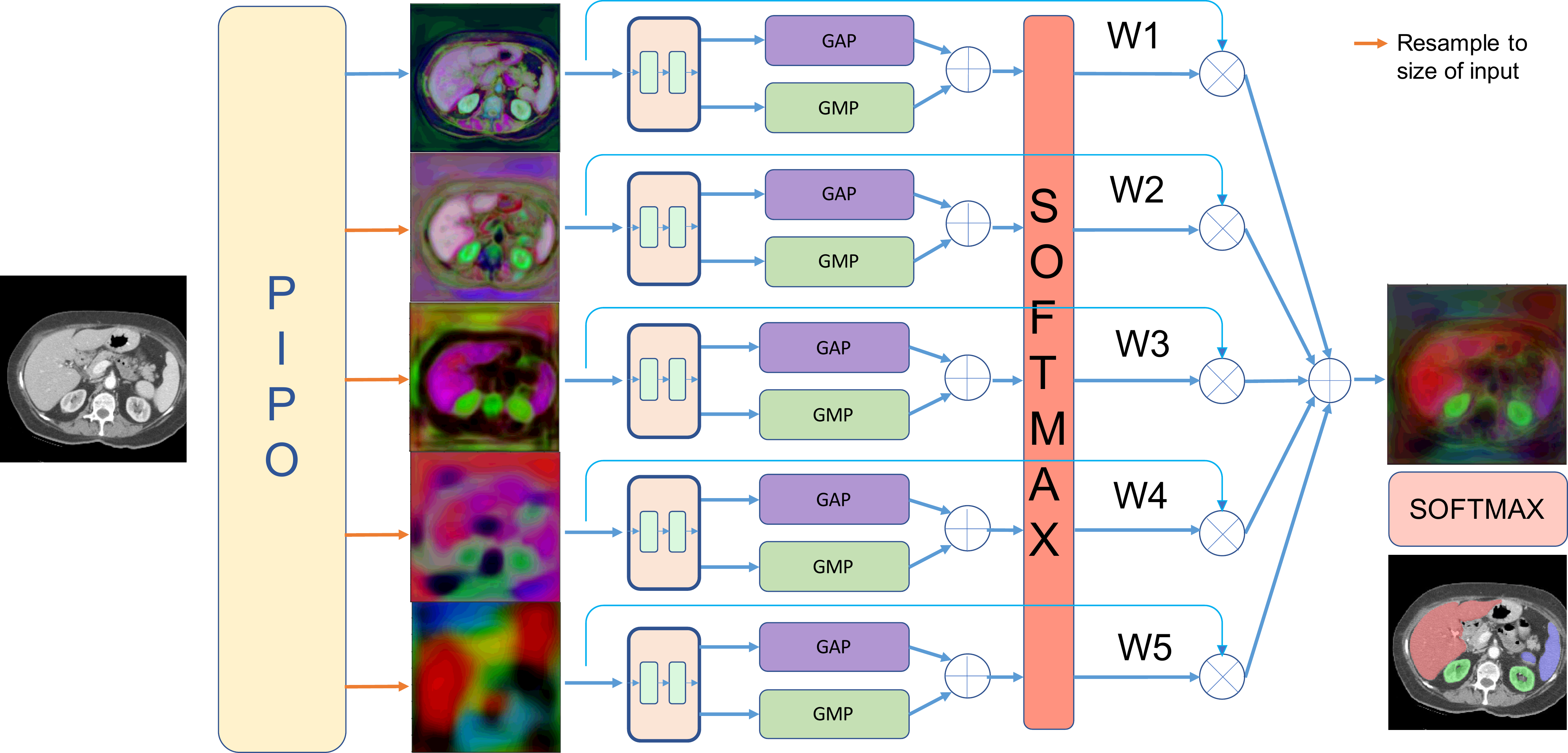}}
	\caption{Adaptive fusion of the multi-scale output segmentation features from PIPO-FAN. Features from lower scales tend to represent specific local segmentation, while features from higher scales are blurry but carry class information.
	Adaptive weights are computed by applying a shared convolutional module to the pyramid output features.
} 
	\label{fig:PIPO_w1}
\end{figure*}

\begin{table*}[t]
\centering
\caption{Comparison of features in U-Net and PIPO-FAN.}
\label{tbl:u_pipo_com}
$\begin{array}{c|c|c}
\toprule
\textbf{Features} & 
\textbf{Encoding} & 
\textbf{Decoding} \\
\midrule
\addlinespace
\textbf{U-Net} & 
  \begin{aligned}
    f_{s,j} = Pool(Conv(f_{s-1,j-1})), \quad \text{$j=s$}
  \end{aligned} &
  \begin{aligned}
    f_{s,j} =
    Concat(Upsample(Conv(f_{s,j-1})), Conv(f_{s,s})), \quad \text{$j=s$}
  \end{aligned} \\
\addlinespace
\midrule
\addlinespace
\textbf{PIPO-FAN} &
  \begin{aligned}
    f_{s,j} = \begin{cases}
    Conv(f_{s,j-1}), & \text{$j<s$}\\
    Conv(f_{s,j-1}) + Pool(Conv(f_{s-1,j-1})), & \text{$j=s$}
    \end{cases}
  \end{aligned} &
  \begin{aligned}
    f_{s,j} = \begin{cases}
    Concat(Upsample(Conv(f_{s,j-1})), Conv(f_{s,s}), & \text{$j+s=2S$}\\
    Conv(f_{s,j-1}), & \text{$j+s>2S$}
    \end{cases}
  \end{aligned} \\
\addlinespace
\bottomrule
\end{array}$
\end{table*}

To seek for patterns from images in different scales, \emph{i.e.} scale invariance, the proposed network first performs pyramid analysis to the input image by using spatial pyramid pooling and shared convolution to obtain context information in different scales. Unlike the classical U-net based methods, where the scale only reduces when the convolutional depth increases, 
PIPO-FAN has multi-scale features at each depth and therefore both global and local context information can be integrated to augment the extracted features. 
After going through one or more convolutional layers, the features are fused together to have hierarchical structural information. The input feature maps to each level of U-Net and PIPO-FAN are listed in Table~\ref{tbl:u_pipo_com} for comparison. In Table~\ref{tbl:u_pipo_com}, $I_s$ denotes the input at the $s$-th scale, $f_{s,j}$ denotes the input features of the $j$-th convolution block at the $s$-th scale, and $S$ is the scale number of inputs. A feature map to the first convolutional block at each scale $f_{s,1}$ is set as $I_s$.

A notable character of PIPO-FAN is that features fused at each level all went through the same number of convolutional layers, i.e. they have equal convolutional depth (ECD). It is achieved by inserting the ResBlocks to the networks, as shown in Fig.~\ref{fig:Concave} by the light blue color boxes. Unlike the existing works, e.g. \cite{mimo_net, fu2018joint}, where features after various depths of convolutions are directly fused together, we deal with the problem of semantic gap using ECD. With the proposed ECD connections, all the fused features at each step are at the same semantic abstraction level to better exploit the pyramid shape of U-Net.

\subsection{Pyramid Output}

Furthermore, inspired by the work of deep supervision \cite{deepsup}, we introduce deep pyramid supervision (DPS) to the decoding path for generating and supervising outputs of different scales. During the training process, we perform spatial pyramid pooling to the ground truth segmentation to generate labels in all output scales. The training loss is computed by using the corresponding output and ground truth segmentation at the same scale. Weighted cross entropy is used as the loss function in our work, which is defined as
\begin{equation} \label{eq:1}
L_{DPS} = -\frac{1}{S}\sum_{s=1}^{S}\frac{1}{N_s}\sum_{i=1}^{N_s}\sum_{c=0}^{3}w_{i,s}^c y_{i,s}^c \log p_{i,s}^c,
\end{equation}
where $p_{i,s}^c$ denotes the predicted probability of voxel $i$ belonging to class $c$ (liver, kidney, spleen or background) in scale $s$
\begin{equation}\label{eq:2}
p_{i,s}^c =  \frac{exp(f_{i,s}^j)}{\sum_{j}^{ }exp(f_{i,s}^j)))},
\end{equation}
$y_{i,s}^c$ is the ground truth label in scale $s$, $N_s$ denotes the number of voxels in the scale $s$, and $w_i^c$ is a weighting parameter for class $c$. 
DPS can help relieve the problem of gradient vanishing in deep neural networks and learn deep level features with hierarchical contexts. It also enforces the outputs in all scales to maintain structural information.

\subsection{Adaptive Fusion}

With the above mentioned DPS mechanism, informative segmentation features are obtained at different scales. Since they may contain complementary context information, we are motivated to fuse these features together to achieve more accurate segmentation. To effectively exploit the contextual information in different scales, we design an adaptive fusion (AF) module to learn the relative importance of each scale and fuse the score maps (i.e., last layer output before softmax) in an automatic fashion. For each pixel of each score map, larger value indicates higher probability belonging to the class. It also implies higher certainty of the prediction. Thus, squeezing one channel of a score map into single core by using a global operation can provide a measure of the prediction confidence.

In particular, this AF module uses an attention mechanism to indicate the importance at each scale. As shown in Fig. \ref{fig:PIPO_w1}, after hierarchical abstraction from pyramid input, pyramid output features are propagated into the attention module. 
To leverage the similar structural information at each scale,
Pyramid outputs are first passed into a shared convolutional block to achieve scale invariance. 
\begin{equation}
F_s = Conv(f_s),  \label{eq:3}
\end{equation}
Those features are then squeezed into a single channel feature vector, which denotes the overall score of output at each scale. Global average pooling (GAP) and global max pooling (GMP) can extract the global certainty score of predictions at each scale. In our work, GAP and GMP are summed up to extract the global score $S_s$ at scale $s$
\begin{equation}
S_s = GAP(F_s) + GMP(F_s). \label{eq:4}
\end{equation}
The values from different scales are then concatenated to feed into a softmax layer to get the corresponding weight for each scale. The weights $w_s$ reflects the importance of feature at scale $s$. After resampling to the original image size, the pyramid output features are summed to get the fused features, where the scale weights are computed as
\begin{equation}
W_s =  \frac{exp(S_s)}{\sum_{s}^{ }exp(S_s)))}.
\end{equation}
Another softmax layer is applied to obtain the final segmentation result
\begin{equation}
p_a^c = \frac{exp(\sum_{s=1}^{S}W_{s}f_{s})}{\sum_{c}^{ }exp(\sum_{s=1}^{S}W_{s}f_{s})))}.
\end{equation}

\subsection{Multi-organ segmentation over multiple datasets}

The proposed target adaptive loss (TAL) allows training a segmentation algorithm over multiple datasets with different labels. To use a cross entropy loss, the model should typically predict the probability distribution for all labels in the dataset as
\begin{equation}
L_{full}  = \sum_{c=0}^{n} y_i^c \log p_i^c.
\end{equation}
Considering the partial labels are subset of all labels, probabilities can be merged according to the known labels. 
Therefore, we treat the unknown labels as background to allow computing the loss. 
TAL function is defined as
\begin{equation}
L_{TAL} = \sum_{c \in C_k} y_i^c \log p_i^c + \mathbf{1}_{[\sum_{c \in C_k}{y_i^{c}} = 0]} \log (1-\sum_{c \in C_k}{p_i^{c}}),
\end{equation}
where $C_k$ denotes the organs labeled in the dataset. $p_i^c$ is the predicted probability of voxel i belonging to class c. The ground truth label $y_i^c$ indicates whether voxel $i$ belonging to class $c$ or not.

TAL can be easily implemented by modifying the last layer of the segmentation network to have multiple branches to segment all organs labeled in these datasets. After the second softmax layer, each branch gets a probability of that class. When training on the partially labeled datasets, probabilities of labeled target organs are preserved, while other probabilities are merged as ``non-target'' class in that dataset. For example, to train on LiTS dataset, we preserve the probability of liver and combine the other branches probabilities to the probability of non-liver. Then a binary cross entropy can be computed as 
\begin{equation}
L_{TAL} = \mathbf{1}_{[y_i^{c1} = 1]} \log p_i^{c1} + \mathbf{1}_{[y_i^{c1} = 0]} \log (1-p_i^{c1}),
\end{equation}
where $c1$ denotes the labeled liver. 
Similarly, training on the KiTS and spleen datasets can be completed. Such a mechanism allows gradients to be  back propagated through the branches with the labeled organs. A general segmentation model can then be obtained, which is able to segment multiple organs when they are present in an image.

\section{Experiments}
\label{sec:experiments}

\begin{figure*}[t]    
	\center{\includegraphics[width = \textwidth] {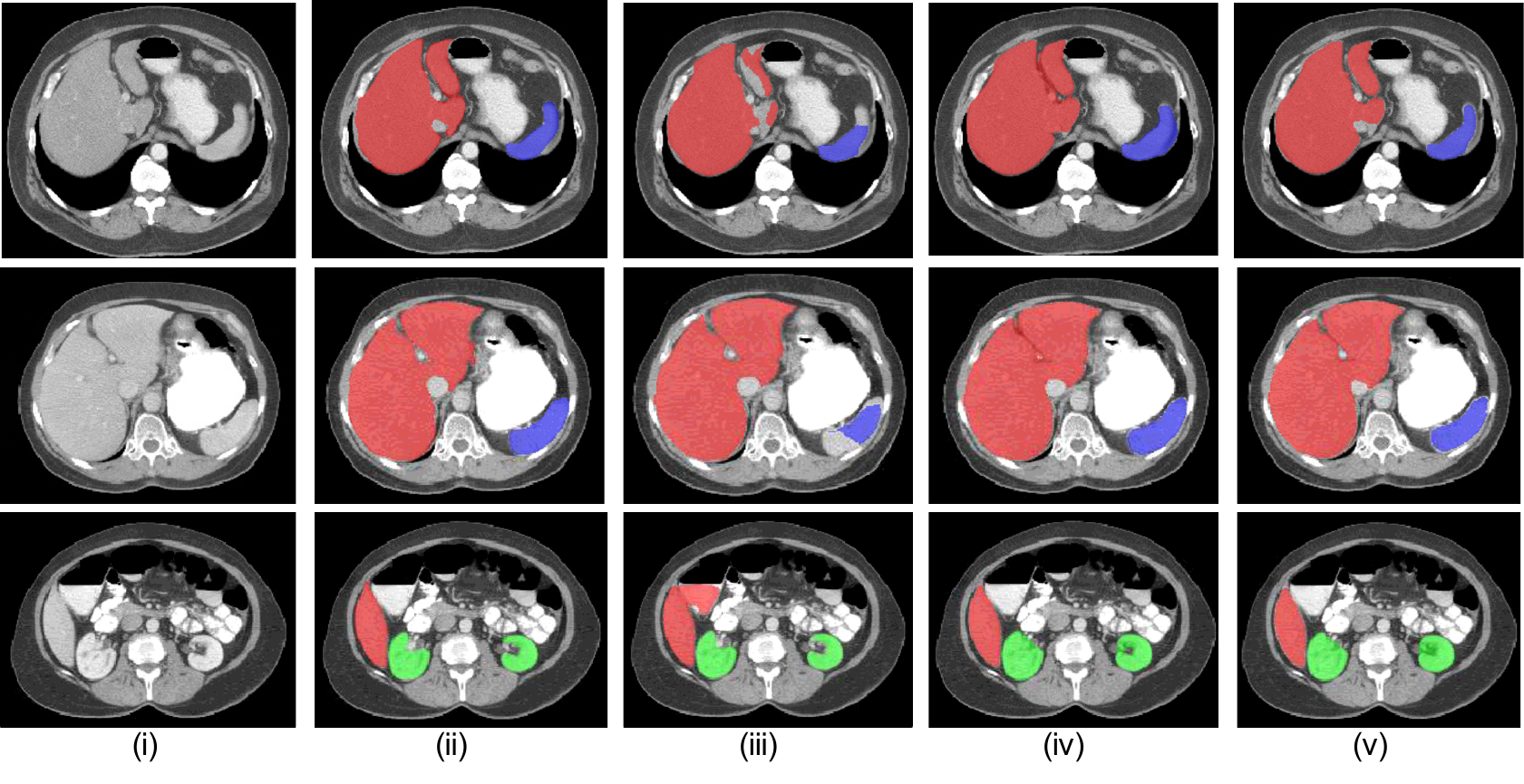}}        
	\caption{Segmentation examples on BTCV data. Each row from left to right shows the following images in order: (i) input image, (ii) ground truth segmentation, (iii) segmentation results of models trained with BTCV only, (iv) with partially labeled data, and (v) with all the datasets, respectively. Red, blue and green colors depict the segmentation of the liver, the spleen, and the kidney, respectively.} 
	\label{fig:seg_examples1}
\end{figure*}

\begin{figure*}[t]    
	\center{\includegraphics[width = \textwidth] {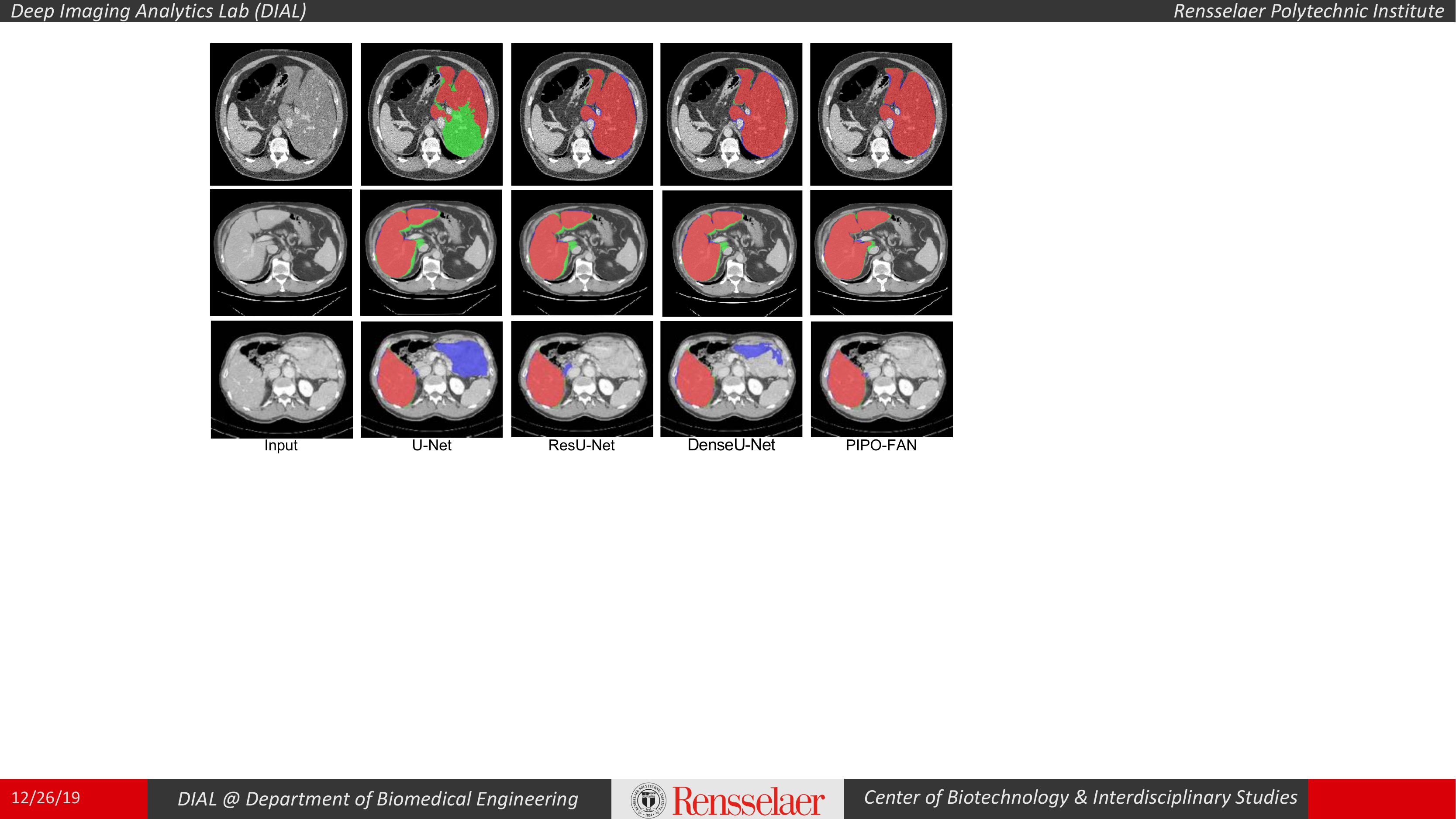}}        
	\caption{Segmentation examples of different methods on LiTS data. From left to right are the raw image, results of U-Net, ResU-Net, DenseU-Net and our proposed PIPO-FAN, the red depicts correctly predicted liver segmentation, the blue shows false positive, green shows false negative.}  
	\label{fig:seg_examples2}
\end{figure*}

\subsection{Materials}

We evaluated our model on four publicly available datasets, LiTS (Liver tumor segmentation challenge) \cite{bilic2019liver}, KiTS (Kidney tumor segmentation challenge) \cite{heller2019kits19}, the Spleen segmentation dataset \cite{simpson2015chemotherapy} and BTCV (Beyond  the  Cranial  Vault) segmentation challenge dataset \cite{ben_2015_btcv}. The first three datasets are single-organ annotation datasets and the last one has multiple organs annotated. Fig.~\ref{fig:data_examples} provides a summary of the available organ annotations in those datasets.

LiTS consists of 131 training and 70 test images. The data were collected from different hospitals and the intra-slice resolutions of the CT scans vary between 0.45mm and 6mm and the inter-slice resolutions vary between 0.6mm and 1.0mm for inter-slices. KiTS consists of 210 training and 90 test images collected from 300 patients, who underwent nephrectomy for kidney tumors. The spleen segmentation dataset is composed of patients undergoing
chemotherapy for liver metastases at Memorial Sloan Kettering Cancer Center (New York, NY, USA) \cite{simpson2015chemotherapy}. The BTCV segmentation challenge dataset contains 47 subjects with segmentation of all abdominal organs except duodenum. 30 of them have labels of both two kidneys. We randomly split those subjects into 21 for training and 9 for validation. We select three abdominal organs in partially labeled datasets as the target organs. 

In these datasets, the size of each slice is 512$\times$512 pixels. To speed up the model training, we resized the axial slices into 256$\times$256 pixels, where the boundary information is still well preserved. We keep the CT imaging HU values using in range of [-200, 200] to have a better contrast on the abdominal organs. Then we normalize all volumetric data into zero mean and unit variance. For each epoch, we randomly select three continuous slices containing target organ label from all the CT training volumes and crop a patch with size of 224$\times$224 as input to the network. After obtaining the segmentation volume, the connected component analysis was performed to keep only the largest component as the segmentation result of liver and spleen, and the largest two components as the segmentation of the kidneys.

\subsection{Implementation Details}

Our implementation is based on the open-source platform PyTorch \cite{paszke2017automatic}. All the convolutional operations are followed by batch normalization and ReLU activation. The channel numbers of feature maps after the convolutional blocks are 64, 128, 256, 512, 512, 512, 256, 128, 64 sequentially. The numbers of input and output scales are both set to be 5. For network training, we use the RMSprop optimizer. For multi-organ segmentation, we set learning rate to be 0.0002 and the maximum number of training epochs to be 4000.  For single organ segmentation, we set the initial learning rate to be 0.002 and the maximum number of training epochs to be 2500. The learning rate decays by 0.01 after every 40 epochs. For the first 2000 epochs, the deep supervision losses are applied to focus on the feature abstraction ability in each scale. For the remaining epochs, the adaptive weighting layer is activated and the deep supervision loss stops, which helps to optimize only the adaptively fused score map output. 

When training a single-organ segmentation model, empirically, we set the weights in Eqn.~(\ref{eq:1}) to be 0.2 and 1.2 for background and the organ, respectively, to counteract the unbalanced training samples. 
When training a multi-organ segmentation model, for simplicity and generalization ability, all the weights in the cross entropy function are set to be 1. When two or more datasets are used, the model is trained alternatively on those datasets by using one at a time.

In our work, five-fold cross validation on LiTS and KiTS was employed to evaluate the performance of the models. Dice score is used as the evaluation criterion. When preparing the test result submission to the LiTS challenge websites, we used majority voting to combine the outputs of the five models to get the final segmentation. 
Our implementation code is open-sourced available at \url{https://github.com/DIAL-RPI/PIPO-FAN}.

\begin{table}[t]
	\centering
	\caption{\label{tab:uni_val} Segmentation performance using different combinations of datasets (Dice \%)}
	\begin{tabular}{lcccc}
		\hline
		
		Datasets &  Liver & Kidney  & Spleen & Average\\
		\hline
		BTCV & 95.8 & 92.7 & 92.3 & 93.6\\
		BTCV + LiTS & 95.6 & 91.6 & 95.4 & 94.2 \\
		BTCV + KiTS & 94.2 & 91.9 & 94.5 & 93.5\\
		BTCV + Spleen & \textbf{95.9} & \textbf{93.5} & 93.8 & \textbf{94.4} \\
		LiTS + KiTS + Spleen & 95.6 & 91.6 & 93.8 & 93.7\\
		All datasets & \textbf{95.9} & 91.9 & \textbf{95.5} & \textbf{94.4}\\
		\hline
	\end{tabular}
\end{table}

\subsection{Multi-organ Segmentation}

We evaluate our proposed training strategy on BTCV validation dataset. Liver, kidney and spleen are three target abdominal organs to compare the segmentation performance. 
In Table~\ref{tab:uni_val}, different combinations of datasets are used to train the multi-organ segmentation model. The proposed PIPO-FAN is used as the segmentation model. Dice score is used as the evaluation criterion. When training on the BTCV dataset, the model serves as the baseline of multi-organ segmentation. Three partially labeled datasets, LiTS, KiTS and Spleen segmentation datasets are individually added as additional training dataset to enhance training procedure.
The combination of four datasets are also used for the unified training. Some example results are shown in Fig.~\ref{fig:seg_examples1}. We noticed that training with the additional partially labeled datasets improves the multi-organ segmentation, especially the spleen dataset. Compared to using BTCV alone, the use of additional datasets significantly boosts the segmentation performance of the spleen, which may be because the spleen has similar appearances across all the datasets.

\begin{table}[t]
	\centering
	\caption{\label{tab:uni_net} Performance comparison with other networks on the BTCV dataset. 
	(Dice \%)}
	\begin{tabular}{lcccc}
		\hline
		
		Architecture & Liver & Kidney & Spleen  &  Average\\
		\hline
		U-Net \cite{unet}   & 95.6 & 89.7 &  91.0 &   92.1 \\
		ResU-Net \cite{han_automatic_2017} & 95.1  & 91.3 & 90.9 & 92.4  \\
		DeepLabV3 \cite{chen2017rethinking} & 94.2 & 86.0 & 87.4 & 89.2\\
		\hline
		PIPO & 95.7 & 92.6 & 90.1  &  92.8  \\
		PIPO-FAN & \textbf{95.8} & \textbf{92.7} & \textbf{92.3} & \textbf{93.6}  \\
		\hline
	\end{tabular}
\end{table}

\begin{table}[t]
	\centering
	\caption{\label{tab:data_net_com2} Performance comparison with other networks on the combined all datasets. 
	(Dice \%)}
	\begin{tabular}{lcccc}
		\hline
		Architecture & Liver & Kidney & Spleen  &  Average\\
		\hline
		U-Net \cite{unet}   & \textbf{95.9} & \textbf{92.7} & 93.5 & 94.0 \\
		DeepLabV3 \cite{chen2017rethinking} & 94.1 & 89.6 & 90.9 & 91.5\\
		PIPO-FAN  & \textbf{95.9} & 91.9 & \textbf{95.5} & \textbf{94.4} \\
		\hline
	\end{tabular}
\end{table}

\subsection{Model Analysis and Ablation Studies}

\begin{table}[t]
	\centering
	\caption{\label{tab:2D_nets} Five-fold cross validation against other benchmark methods on two open challenge datasets. (Dice \%)}
	\begin{tabular}{lcc}
		\hline
		
		Architecture & LiTS & KiTS  \\
		\hline
		U-Net \cite{unet}   & 93.9 $\pm$ 0.50 &  95.8 $\pm$ 0.91 \\
		ResU-Net \cite{han_automatic_2017} & 94.1 $\pm$ 0.88   & 94.8 $\pm$ 1.06  \\
		DenseU-Net \cite{li_hdenseu} &  94.1 $\pm$ 0.30   & 94.2 $\pm$ 2.08 \\
		\hline
		PIPO & 95.3 $\pm$ 0.62 &  \textbf{96.5 $\pm$ 0.55} \\
		PIPO-FAN & \textbf{95.6 $\pm$ 0.48}  & 96.2 $\pm$ 1.02  \\
		\hline
	\end{tabular}
\end{table}

We further compared our proposed PIPO-FAN against several other classical benchmark 2D segmentation networks, including U-Net \cite{unet}, ResU-Net \cite{han_automatic_2017}, DenseU-Net \cite{li_hdenseu} and DeepLabV3~\cite{chen2017rethinking} to demonstrate the effectiveness of DPS and AF. For fair comparison, U-Net, ResU-Net and our PIPO-FAN are all 19-layer networks. DeepLabV3 uses ResNet18~\cite{He2016DeepRL} as its backbone.
The segmentation results on BTCV are shown in Table~\ref{tab:uni_net}.
For comparison on partially labeled multi-organ segmentation, the results on the combined all datasets are shown in Table~\ref{tab:data_net_com2}. PIPO-FAN outperforms all other networks on multi-organ segmentation in both the BTCV dataset and the combined all datasets. Some example results on LiTS are shown in Fig.~\ref{fig:seg_examples2}. 
The DenseU-Net is the 2D DenseU-Net architecture used in \cite{li_hdenseu} and the encoder part is Densenet-169. According to the experimental results from \cite{heller2019state}, adding residual connections makes little difference in performance on the KiTS data. Thus, in our experiments on the KiTS data, we used ConvBlocks in PIPO-FAN rather than ResBlocks. All these 2D networks are trained from scratch in the same environment. 
We evaluate the performance of the above networks on each dataset through five-fold cross validation. The 131 labeled LiTS data and 210 KiTS data are split into 5 folds, respectively. Each fold is used only once for validation, while the other four are used for training. 

The five-fold cross validation results are shown in Table~\ref{tab:2D_nets}.
The conducted $t$-test shows that PIPO-FAN significantly outperforms U-Net, ResU-Net and DenseU-Net with $p$-values of 0.004, 0.025, and 0.002 in LiTS data respectively.

\begin{table}[t]
\centering
\caption{\label{tab:val} Ablation study of PIPO-FAN network structures on LiTS dataset (Dice \%)}
	\begin{tabular}{lcc}
		\hline
		
		Architecture &  Avg. Dice & Glb. Dice\\
		\hline
		Single-scale input/output & 94.1 & 94.5 \\
PIPO w/o ECD & 95.1 & 95.2\\
PIPO-FAN w/o ECD & 95.2 & 95.1\\
PIPO with ECD & 95.3 & 95.4\\
PIPO-FAN with ECD & \textbf{95.6} & \textbf{95.8}\\
		\hline
	\end{tabular}
\end{table}

We first evaluated the effectiveness of the proposed equal convolutional depth (ECD) mechanism presented in Section~\ref{sec:ECD}. Table~\ref{tab:val} shows the the segmentation performance under different network configurations. As expected, using PIPO always outperforms single-scale input/output segmentation, which is indeed a ResU-Net. With PIPO to explore multi-scale image information, using ECD results in consistent performance improvement. It is worth noting that using ECD with only deep pyramid supervision (DPS) performs better than using AF without ECD. This not only shows that ECD is effective in extracting image features, but also illustrates the necessity of having good features for adaptive fusion module to work efficiently.

\begin{table}[t]
\centering
\caption{\label{tab:test} Performance evaluation of varying the numbers of the input and output scales on LiTS dataset.}
\begin{tabular}{cccc}
\hline
Input scale & Output scale & Avg. Dice & Parameter numbers\\
 \hline
  1 & 1     & 94.1 & 13,395,394 \\
  3 & 1     &   94.9 & 15,981,634  \\
  5 & 1    &   94.5 & 24,740,034 \\
  3 & 3     &  95.1 & 16,258,566 \\
  5 & 3    & 95.3 & 25,016,966 \\
  5 & 5    & \textbf{95.6} & 28,270,986 \\
 \hline
\end{tabular}
\end{table}

We also tried different number of scales for the input and output to evaluate the relationship between scales and model capacity. The results are shown in Table~\ref{tab:test}. It can be seen that the larger scale numbers of input scales and output scales, the better segmentation accuracy the model obtains. It may be because the higher scales can provide larger receptive fields and thus enhance the contextual information, This in turn helps the feature abstraction, i.e. extracting representative segmentation features. A special case is that when the number of input scale increased from 3 to 5, but the number of output scale remains at 1, the segmentation performance dropped. It may be because increasing the input scale alone without additional output supervision adds difficulty to the network training.  
Based on the results, we empirically make use of five input and output scales in the final version of our work.

\subsection{Comparison with state-of-the-arts on LiTS challenge}

\begin{table}[t]
\centering
\caption{\label{tab:comparison} Comparison of segmentation accuracy (Dice \%) on the LiTS test dataset. Results are taken from the challenge website (accessed on September 11, 2019).}
\begin{tabular}{lccc }
\hline
Methods & \# of Steps & Avg. Dice & Glb. Dice\\
 \hline
Vorontsov et al. \cite{voron_liver} & 1 & 95.1    & -  \\
H-DenseUNet \cite{li_hdenseu}   & 2 & 96.1    & 96.5 \\
DeepX \cite{yuan2017hierarchical} & 2 & 96.3  & 96.7  \\
2D DenseUNet \cite{li_hdenseu}   & 2 & 95.3    & 95.9 \\
PIPO-FAN (ours)   & 1 & 96.1 & 96.5 \\
 \hline
\end{tabular}
\end{table}

Most of the state-of-the-art methods on liver CT image segmentation takes two steps to complete the task, where a coarse segmentation is used to locate the liver followed by fine segmentation step to obtain the final segmentation \cite{han_automatic_2017,li_hdenseu}. 
However, such two-step methods can be computationally expensive. For example, the method in \cite{li_hdenseu} takes 21 hours to finetune a pretrained 2D DenseUNet and another 9 hours to finetune the H-DenseUNet with two Titan Xp GPUs. In contrast, our proposed method can be trained on a single Titan Xp GPU in 3 hours. More importantly, when segmenting a CT volume, our method only takes 0.04s for one slice on a single GPU, which is, to the best of our knowledge, the fastest segmentation method compared to other reported methods. In the same, we are able to obtain the same Dice performance and even better symmetric surface distance (SSD) (ASSD: 1.413 $<$ 1.450, MSSD: 24.408 $<$ 27.118, 2.421 $<$ 3.150). Table~\ref{tab:comparison} shows the performance comparison with other published state-of-the-art methods on LiTS challenge dataset. Despite its simplicity, our proposed 2D network segments the liver in a single step and can obtain a very competitive performance with less than 0.2\% drop in Dice, compared to the top performing method -- DeepX \cite{yuan2017hierarchical}. 

\section{Discussion and Conclusion}
In this work, the proposed method provides a universal training strategy for better utilization of partially annotated datasets for multi-organ segmentation. Robust and accurate multi-organ segmentation can help improve clinical applications such as computer-aided diagnosis and surgery. Benefited from the flexibility in annotations for training, the work can be extended to train networks on heterogeneous datasets focusing on different regions (e.g. abdominal, chest, pelvic, and whole body). That makes it possible to use one network to segment different CT images. To leverage the information from large scale datasets, previous works presented semi-supervised methods use weakly labeled and even unlabeled data. Our work is complementary to those methods and can be combined with semi-supervised learning to further alleviate the problem of data hunger in multi-organ segmentation.

In summary, a novel network architecture for multi-scale feature abstraction, which incorporates multi-scale features in a hierarchical fashion at various depths for medical image segmentation, is presented in this paper.
The proposed 2D network with only a single step shows very competitive performance compared with other multi-step 3D networks in CT image segmentation. 
We further develop a unified segmentation strategy to train the proposed network on multiple partially labeled datasets for multi-organ segmentation. The new strategy gives the segmentation network better robustness and accuracy by enlarging the training dataset. The source code of our work has been open sourced to enable further testing and development in a larger scale on other imaging modalities.

\section*{Acknowledgment}

The authors would like to thank NVIDIA Corporation for the donation of two Titan Xp GPUs used for this research. We would also like to thank Prof. George Xu (RPI), Mr. Zhao Peng (USTC), and Dr. Sheng Xu (NIH) for the insightful discussions.

\bibliographystyle{IEEEtran}
\bibliography{refs}

\begin{thebibliography}{10}
\providecommand{\url}[1]{#1}
\csname url@samestyle\endcsname
\providecommand{\newblock}{\relax}
\providecommand{\bibinfo}[2]{#2}
\providecommand{\BIBentrySTDinterwordspacing}{\spaceskip=0pt\relax}
\providecommand{\BIBentryALTinterwordstretchfactor}{4}
\providecommand{\BIBentryALTinterwordspacing}{\spaceskip=\fontdimen2\font plus
\BIBentryALTinterwordstretchfactor\fontdimen3\font minus
  \fontdimen4\font\relax}
\providecommand{\BIBforeignlanguage}[2]{{%
\expandafter\ifx\csname l@#1\endcsname\relax
\typeout{** WARNING: IEEEtran.bst: No hyphenation pattern has been}%
\typeout{** loaded for the language `#1'. Using the pattern for}%
\typeout{** the default language instead.}%
\else
\language=\csname l@#1\endcsname
\fi
#2}}
\providecommand{\BIBdecl}{\relax}
\BIBdecl

\bibitem{rueckert_atlas_seg_2013}
R.~{Wolz}, C.~{Chu}, K.~{Misawa}, M.~{Fujiwara}, K.~{Mori}, and D.~{Rueckert},
  ``Automated abdominal multi-organ segmentation with subject-specific atlas
  generation,'' \emph{IEEE Transactions on Medical Imaging}, vol.~32, no.~9,
  pp. 1723--1730, Sep. 2013.

\bibitem{li_hdenseu}
X.~{Li}, H.~{Chen}, X.~{Qi}, Q.~{Dou}, C.~{Fu}, and P.~{Heng}, ``H-denseunet:
  Hybrid densely connected unet for liver and tumor segmentation from {CT}
  volumes,'' \emph{IEEE Transactions on Medical Imaging}, vol.~37, no.~12, pp.
  2663--2674, Dec 2018.

\bibitem{zhu_boundary}
Q.~{Zhu}, B.~{Du}, and P.~{Yan}, ``Boundary-weighted domain adaptive neural
  network for prostate mr image segmentation,'' \emph{IEEE Transactions on
  Medical Imaging}, pp. 1--1, 2019.

\bibitem{8451958}
K.~{He}, X.~{Cao}, Y.~{Shi}, D.~{Nie}, Y.~{Gao}, and D.~{Shen}, ``Pelvic organ
  segmentation using distinctive curve guided fully convolutional networks,''
  \emph{IEEE Transactions on Medical Imaging}, vol.~38, no.~2, pp. 585--595,
  Feb 2019.

\bibitem{DR_drinet}
L.~{Chen}, P.~{Bentley}, K.~{Mori}, K.~{Misawa}, M.~{Fujiwara}, and
  D.~{Rueckert}, ``Drinet for medical image segmentation,'' \emph{IEEE
  Transactions on Medical Imaging}, vol.~37, no.~11, pp. 2453--2462, Nov 2018.

\bibitem{LongSD15}
\BIBentryALTinterwordspacing
J.~Long, E.~Shelhamer, and T.~Darrell, ``Fully convolutional networks for
  semantic segmentation,'' in \emph{{CVPR}}, 2015, pp. 3431--3440. [Online].
  Available: \url{https://doi.org/10.1109/CVPR.2015.7298965}
\BIBentrySTDinterwordspacing

\bibitem{unet}
O.~Ronneberger, P.~Fischer, and T.~Brox, ``U-net: Convolutional networks for
  biomedical image segmentation,'' in \emph{{MICCAI}}, 2015, pp. 234--241.

\bibitem{Chen2015AttentionTS}
L.-C. Chen, Y.~Yang, J.~Wang, W.~Xu, and A.~L. Yuille, ``Attention to scale:
  Scale-aware semantic image segmentation,'' \emph{2016 IEEE Conference on
  Computer Vision and Pattern Recognition (CVPR)}, pp. 3640--3649, 2015.

\bibitem{He2016DeepRL}
K.~He, X.~Zhang, S.~Ren, and J.~Sun, ``Deep residual learning for image
  recognition,'' in \emph{{CVPR}}, 2016, pp. 770--778.

\bibitem{Huang2017DenselyCC}
G.~Huang, Z.~Liu, L.~van~der Maaten, and K.~Q. Weinberger, ``Densely connected
  convolutional networks,'' in \emph{{CVPR}}, 2017, pp. 2261--2269.

\bibitem{han_automatic_2017}
\BIBentryALTinterwordspacing
X.~Han, ``Automatic {Liver} {Lesion} {Segmentation} {Using} {A} {Deep}
  {Convolutional} {Neural} {Network} {Method},'' \emph{arXiv: 1704.07239}, Apr.
  2017. [Online]. Available: \url{http://arxiv.org/abs/1704.07239}
\BIBentrySTDinterwordspacing

\bibitem{lin2016efficient}
G.~Lin, C.~Shen, A.~Van Den~Hengel, and I.~Reid, ``Efficient piecewise training
  of deep structured models for semantic segmentation,'' in \emph{{CVPR}},
  2016, pp. 3194--3203.

\bibitem{psp}
\BIBentryALTinterwordspacing
H.~Zhao, J.~Shi, X.~Qi, X.~Wang, and J.~Jia, ``Pyramid scene parsing network,''
  in \emph{{CVPR}}.\hskip 1em plus 0.5em minus 0.4em\relax IEEE Computer
  Society, July 2017, pp. 6230--6239. [Online]. Available:
  \url{https://doi.ieeecomputersociety.org/10.1109/CVPR.2017.660}
\BIBentrySTDinterwordspacing

\bibitem{fang2019unified}
X.~Fang, B.~Du, S.~Xu, B.~J. Wood, and P.~Yan, ``Unified multi-scale feature
  abstraction for medical image segmentation,'' \emph{arXiv preprint
  arXiv:1910.11456}, 2019.

\bibitem{kamnitsas2017efficient}
K.~Kamnitsas, C.~Ledig, V.~F. Newcombe, J.~P. Simpson, A.~D. Kane, D.~K. Menon
  \emph{et~al.}, ``Efficient multi-scale {3D CNN} with fully connected {CRF}
  for accurate brain lesion segmentation,'' \emph{Medical Image Analysis},
  vol.~36, pp. 61--78, 2017.

\bibitem{simpson2015chemotherapy}
A.~L. Simpson, J.~N. Leal, A.~Pugalenthi, P.~J. Allen, R.~P. DeMatteo, Y.~Fong
  \emph{et~al.}, ``Chemotherapy-induced splenic volume increase is
  independently associated with major complications after hepatic resection for
  metastatic colorectal cancer,'' \emph{Journal of the American College of
  Surgeons}, vol. 220, no.~3, pp. 271--280, 2015.

\bibitem{liu2020deep}
L.~Liu, W.~Ouyang, X.~Wang, P.~Fieguth, J.~Chen, X.~Liu, and
  M.~Pietik{\"a}inen, ``Deep learning for generic object detection: A survey,''
  \emph{International Journal of Computer Vision}, vol. 128, no.~2, pp.
  261--318, 2020.

\bibitem{dong2018holistic}
S.~Dong, Z.~Gao, S.~Sun, X.~Wang, M.~Li, H.~Zhang, G.~Yang, H.~Liu, and S.~Li,
  ``Holistic and deep feature pyramids for saliency detection,'' in \emph{29TH
  BRITISH MACHINE VISION CONFERENCE (BMVC)}, 2018.

\bibitem{xueying_isbi}
X.~{Chen}, R.~{Zhang}, and P.~{Yan}, ``Feature fusion encoder decoder network
  for automatic liver lesion segmentation,'' in \emph{2019 IEEE 16th
  International Symposium on Biomedical Imaging (ISBI 2019)}, April 2019, pp.
  430--433.

\bibitem{lin2017feature}
T.-Y. Lin, P.~Doll{\'a}r, R.~Girshick, K.~He, B.~Hariharan, and S.~Belongie,
  ``Feature pyramid networks for object detection,'' in \emph{Proceedings of
  the IEEE conference on computer vision and pattern recognition}, 2017, pp.
  2117--2125.

\bibitem{oktay2018attention}
O.~Oktay, J.~Schlemper, L.~L. Folgoc, M.~Lee, M.~Heinrich, K.~Misawa
  \emph{et~al.}, ``Attention u-net: Learning where to look for the pancreas,''
  \emph{arXiv preprint arXiv:1804.03999}, 2018.

\bibitem{10.1007/978-3-030-00889-5_1}
Z.~Zhou, M.~M. Rahman~Siddiquee, N.~Tajbakhsh, and J.~Liang, ``Unet++: A nested
  u-net architecture for medical image segmentation,'' in \emph{DL in Medical
  Image Analysis and Multimodal Learning for Clinical Decision Support}, 2018,
  pp. 3--11.

\bibitem{deepsup}
C.-Y. Lee, S.~Xie, P.~Gallagher, Z.~Zhang, and Z.~Tu, ``{Deeply-Supervised
  Nets},'' in \emph{Int. Conf. Artificial Intelligence and Statistics}, 09--12
  May 2015, pp. 562--570.

\bibitem{lalonde2018capsules}
R.~LaLonde and U.~Bagci, ``Capsules for object segmentation,'' \emph{arXiv
  preprint arXiv:1804.04241}, 2018.

\bibitem{DBLP:journals/pami/ChenPKMY18}
\BIBentryALTinterwordspacing
L.~Chen, G.~Papandreou, I.~Kokkinos, K.~Murphy, and A.~L. Yuille, ``Deeplab:
  Semantic image segmentation with deep convolutional nets, atrous convolution,
  and fully connected crfs,'' \emph{{IEEE} Trans. PAMI}, vol.~40, no.~4, pp.
  834--848, 2018. [Online]. Available:
  \url{https://doi.org/10.1109/TPAMI.2017.2699184}
\BIBentrySTDinterwordspacing

\bibitem{8662594}
Z.~{Gu}, J.~{Cheng}, H.~{Fu}, K.~{Zhou}, H.~{Hao}, Y.~{Zhao} \emph{et~al.},
  ``Ce-net: Context encoder network for 2d medical image segmentation,''
  \emph{IEEE Transactions on Medical Imaging}, vol.~38, no.~10, pp. 2281--2292,
  Oct 2019.

\bibitem{qin2018autofocus}
Y.~Qin, K.~Kamnitsas, S.~Ancha, J.~Nanavati, G.~Cottrell, A.~Criminisi
  \emph{et~al.}, ``Autofocus layer for semantic segmentation,'' in
  \emph{International Conference on Medical Image Computing and Computer
  Assisted Intervention (MICCAI)}, September 2018.

\bibitem{hier_lecun}
C.~{Farabet}, C.~{Couprie}, L.~{Najman}, and Y.~{LeCun}, ``Learning
  hierarchical features for scene labeling,'' \emph{IEEE Transactions on
  Pattern Analysis and Machine Intelligence}, vol.~35, no.~8, pp. 1915--1929,
  Aug 2013.

\bibitem{mimo_net}
S.~E.~A. {Raza}, L.~{Cheung}, D.~{Epstein}, S.~{Pelengaris}, M.~{Khan}, and
  N.~M. {Rajpoot}, ``{MIMO-Net}: A multi-input multi-output convolutional
  neural network for cell segmentation in fluorescence microscopy images,'' in
  \emph{IEEE 14th International Symposium on Biomedical Imaging (ISBI)}, April
  2017, pp. 337--340.

\bibitem{fu2018joint}
H.~Fu, J.~Cheng, Y.~Xu, D.~W.~K. Wong, J.~Liu, and X.~Cao, ``Joint optic disc
  and cup segmentation based on multi-label deep network and polar
  transformation,'' \emph{IEEE transactions on medical imaging}, vol.~37,
  no.~7, pp. 1597--1605, 2018.

\bibitem{rundo2019use}
L.~Rundo, C.~Han, Y.~Nagano, J.~Zhang, R.~Hataya, C.~Militello \emph{et~al.},
  ``{USE-Net}: incorporating squeeze-and-excitation blocks into u-net for
  prostate zonal segmentation of multi-institutional mri datasets,''
  \emph{arXiv preprint arXiv:1904.08254}, 2019.

\bibitem{peng2019method}
Z.~Peng, X.~Fang, P.~Yan, H.~Shan, T.~Liu, X.~Pei \emph{et~al.}, ``A method of
  rapid quantification of patient-specific organ dose for ct using coupled deep
  multi-organ segmentation algorithms and gpu-accelerated monte carlo dose
  computing code,'' \emph{arXiv preprint arXiv:1908.00360}, 2019.

\bibitem{hong2016learning}
S.~Hong, J.~Oh, H.~Lee, and B.~Han, ``Learning transferrable knowledge for
  semantic segmentation with deep convolutional neural network,'' in
  \emph{Proceedings of the IEEE Conference on Computer Vision and Pattern
  Recognition}, 2016, pp. 3204--3212.

\bibitem{papandreou2015weakly}
G.~Papandreou, L.-C. Chen, K.~P. Murphy, and A.~L. Yuille, ``Weakly-and
  semi-supervised learning of a deep convolutional network for semantic image
  segmentation,'' in \emph{Proceedings of the IEEE international conference on
  computer vision}, 2015, pp. 1742--1750.

\bibitem{meletis_heter_2018}
P.~{Meletis} and G.~{Dubbelman}, ``Training of convolutional networks on
  multiple heterogeneous datasets for street scene semantic segmentation,'' in
  \emph{2018 IEEE Intelligent Vehicles Symposium (IV)}, June 2018, pp.
  1045--1050.

\bibitem{kong_heseg_2019}
F.~{Kong}, C.~{Chen}, B.~{Huang}, L.~M. {Collins}, K.~{Bradbury}, and J.~M.
  {Malof}, ``Training a single multi-class convolutional segmentation network
  using multiple datasets with heterogeneous labels: preliminary results,'' in
  \emph{IGARSS 2019 - 2019 IEEE International Geoscience and Remote Sensing
  Symposium}, July 2019, pp. 3903--3906.

\bibitem{tang2019improving}
Y.~Tang, Y.~Huo, Y.~Xiong, H.~Moon, A.~Assad, T.~K. Moyo, M.~R. Savona,
  R.~Abramson, and B.~A. Landman, ``Improving splenomegaly segmentation by
  learning from heterogeneous multi-source labels,'' in \emph{Medical Imaging
  2019: Image Processing}, vol. 10949.\hskip 1em plus 0.5em minus 0.4em\relax
  International Society for Optics and Photonics, 2019, p. 1094908.

\bibitem{cerrolaza2015automatic}
J.~J. Cerrolaza, M.~Reyes, R.~M. Summers, M.~{\'A}. Gonz{\'a}lez-Ballester, and
  M.~G. Linguraru, ``Automatic multi-resolution shape modeling of multi-organ
  structures,'' \emph{Medical image analysis}, vol.~25, no.~1, pp. 11--21,
  2015.

\bibitem{okada2015abdominal}
T.~Okada, M.~G. Linguraru, M.~Hori, R.~M. Summers, N.~Tomiyama, and Y.~Sato,
  ``Abdominal multi-organ segmentation from ct images using conditional
  shape--location and unsupervised intensity priors,'' \emph{Medical image
  analysis}, vol.~26, no.~1, pp. 1--18, 2015.

\bibitem{robin_2012_miccai}
R.~Wolz, C.~Chu, K.~Misawa, K.~Mori, and D.~Rueckert, ``Multi-organ abdominal
  ct segmentation using hierarchically weighted subject-specific atlases,'' in
  \emph{Medical Image Computing and Computer-Assisted Intervention -- MICCAI
  2012}, N.~Ayache, H.~Delingette, P.~Golland, and K.~Mori, Eds.\hskip 1em plus
  0.5em minus 0.4em\relax Berlin, Heidelberg: Springer Berlin Heidelberg, 2012,
  pp. 10--17.

\bibitem{tong2015discriminative}
T.~Tong, R.~Wolz, Z.~Wang, Q.~Gao, K.~Misawa, M.~Fujiwara \emph{et~al.},
  ``Discriminative dictionary learning for abdominal multi-organ
  segmentation,'' \emph{Medical image analysis}, vol.~23, no.~1, pp. 92--104,
  2015.

\bibitem{herve_2014_miccai}
H.~Lombaert, D.~Zikic, A.~Criminisi, and N.~Ayache, ``Laplacian forests:
  Semantic image segmentation by guided bagging,'' in \emph{Medical Image
  Computing and Computer-Assisted Intervention -- MICCAI 2014}, P.~Golland,
  N.~Hata, C.~Barillot, J.~Hornegger, and R.~Howe, Eds.\hskip 1em plus 0.5em
  minus 0.4em\relax Cham: Springer International Publishing, 2014, pp.
  496--504.

\bibitem{zografos2015hierarchical}
V.~Zografos, A.~Valentinitsch, M.~Rempfler, F.~Tombari, and B.~Menze,
  ``Hierarchical multi-organ segmentation without registration in 3d abdominal
  ct images,'' in \emph{International MICCAI Workshop on Medical Computer
  Vision}.\hskip 1em plus 0.5em minus 0.4em\relax Springer, 2015, pp. 37--46.

\bibitem{eli_2018_deepvnet}
E.~{Gibson}, F.~{Giganti}, Y.~{Hu}, E.~{Bonmati}, S.~{Bandula}, K.~{Gurusamy}
  \emph{et~al.}, ``Automatic multi-organ segmentation on abdominal ct with
  dense v-networks,'' \emph{IEEE Transactions on Medical Imaging}, vol.~37,
  no.~8, pp. 1822--1834, Aug 2018.

\bibitem{chen2017towards}
S.~Chen, H.~Roth, S.~Dorn, M.~May, A.~Cavallaro, M.~M. Lell \emph{et~al.},
  ``Towards automatic abdominal multi-organ segmentation in dual energy ct
  using cascaded 3d fully convolutional network,'' \emph{arXiv preprint
  arXiv:1710.05379}, 2017.

\bibitem{roth2018application}
H.~R. Roth, H.~Oda, X.~Zhou, N.~Shimizu, Y.~Yang, Y.~Hayashi \emph{et~al.},
  ``An application of cascaded 3d fully convolutional networks for medical
  image segmentation,'' \emph{Computerized Medical Imaging and Graphics},
  vol.~66, pp. 90--99, 2018.

\bibitem{wang2019abdominal}
Y.~Wang, Y.~Zhou, W.~Shen, S.~Park, E.~K. Fishman, and A.~L. Yuille,
  ``Abdominal multi-organ segmentation with organ-attention networks and
  statistical fusion,'' \emph{Medical image analysis}, vol.~55, pp. 88--102,
  2019.

\bibitem{Durand_2019_CVPR}
T.~Durand, N.~Mehrasa, and G.~Mori, ``Learning a deep convnet for multi-label
  classification with partial labels,'' in \emph{The IEEE Conference on
  Computer Vision and Pattern Recognition (CVPR)}, June 2019.

\bibitem{bilic2019liver}
P.~Bilic, P.~F. Christ, E.~Vorontsov, G.~Chlebus, H.~Chen, Q.~Dou
  \emph{et~al.}, ``The liver tumor segmentation benchmark (lits),'' \emph{arXiv
  preprint arXiv:1901.04056}, 2019.

\bibitem{heller2019kits19}
N.~Heller, N.~Sathianathen, A.~Kalapara, E.~Walczak, K.~Moore, H.~Kaluzniak
  \emph{et~al.}, ``The {KiTS19} challenge data: 300 kidney tumor cases with
  clinical context, ct semantic segmentations, and surgical outcomes,''
  \emph{arXiv preprint arXiv:1904.00445}, 2019.

\bibitem{ben_2015_btcv}
B.~Landman, Z.~Xu, J.~E. Igelsias, M.~Styner, T.~R. Langerak, and A.~Klein,
  ``{MICCAI} multi-atlas labeling beyond the cranial vault -- workshop and
  challenge,'' 2015.

\bibitem{paszke2017automatic}
A.~Paszke, S.~Gross, S.~Chintala, G.~Chanan, E.~Yang, Z.~DeVito \emph{et~al.},
  ``Automatic differentiation in {PyTorch},'' in \emph{NIPS 2017 Workshop
  Autodiff}, 2017.

\bibitem{chen2017rethinking}
L.-C. Chen, G.~Papandreou, F.~Schroff, and H.~Adam, ``Rethinking atrous
  convolution for semantic image segmentation,'' \emph{arXiv preprint
  arXiv:1706.05587}, 2017.

\bibitem{heller2019state}
N.~Heller, F.~Isensee, K.~H. Maier-Hein, X.~Hou, C.~Xie, F.~Li \emph{et~al.},
  ``The state of the art in kidney and kidney tumor segmentation in
  contrast-enhanced ct imaging: Results of the {KiTS19} challenge,''
  \emph{arXiv preprint arXiv:1912.01054}, 2019.

\bibitem{voron_liver}
E.~{Vorontsov}, A.~{Tang}, C.~{Pal}, and S.~{Kadoury}, ``Liver lesion
  segmentation informed by joint liver segmentation,'' in \emph{{ISBI}}, April
  2018, pp. 1332--1335.

\bibitem{yuan2017hierarchical}
Y.~Yuan, ``Hierarchical convolutional-deconvolutional neural networks for
  automatic liver and tumor segmentation,'' \emph{arXiv:1710.04540}, 2017.

\end{thebibliography}

\end{document}